\documentclass[journal,twoside,web]{ieeecolor}

\newif\ifrevision
\revisionfalse  

\ifrevision
    \newcommand{\revise}[1]{\textcolor{blue}{#1}} 
\else
    \newcommand{\revise}[1]{#1} 
\fi

\usepackage{tmi}
\usepackage{cite}
\usepackage{amsmath,amssymb,amsfonts}
\usepackage{algorithmic}
\usepackage{graphicx}
\usepackage{textcomp}

\usepackage{bbm}
\usepackage{url}
\usepackage{booktabs}
\usepackage{multirow}
\usepackage{pifont}
\let\labelindent\relax
\usepackage{enumitem}

\usepackage{array}
\newcolumntype{C}[1]{>{\centering\arraybackslash}m{#1}}

\def\BibTeX{{\rm B\kern-.05em{\sc i\kern-.025em b}\kern-.08em
    T\kern-.1667em\lower.7ex\hbox{E}\kern-.125emX}}
\begin{document}
\title{Leveraging Segment Anything Model for Source-Free Domain Adaptation via Dual Feature Guided Auto-Prompting}
\author{Zheang Huai, Hui Tang, Yi Li, Zhuangzhuang Chen, and Xiaomeng Li, \IEEEmembership{Member, IEEE}
\thanks{The work described in this paper was supported by a grant from the Research Grants Council (RGC) of the Hong Kong Special Administrative Region, China (Project No. R6005-24) and a research grant from the RGC under the Hong Kong Joint Research Scheme (JRS) of the National Natural Science Foundation of China (NSFC)/RGC, Project No. N\_HKUST654/24. Zheang Huai, Hui Tang, Yi Li, Zhuangzhuang Chen,  and Xiaomeng Li are with the Department of Electronic and Computer
Engineering, The Hong Kong University of Science and Technology,
Hong Kong, China (e-mail: zhuaiaa@connect.ust.hk; eehtang@ust.hk; ylini@connect.ust.hk; eezzchen@ust.hk; eexmli@ust.hk).}
\thanks{Xiaomeng Li is the corresponding author.}
\vspace{-2.0em}
}

\maketitle

\begin{abstract}
Source-free domain adaptation (SFDA) for segmentation aims at adapting a model trained in the source domain to perform well in the target domain with only the source model and unlabeled target data.
Inspired by the recent success of Segment Anything Model (SAM) which exhibits the generality of segmenting images of various modalities and in different domains given human-annotated prompts like bounding boxes or points, we for the first time explore the potentials of Segment Anything Model for SFDA via automatedly finding an accurate bounding box prompt. We find that the bounding boxes directly generated with existing SFDA approaches are defective due to the domain gap.
To tackle this issue, we propose a novel Dual Feature Guided (DFG) auto-prompting approach to search for the box prompt. Specifically, the source model is first trained in a feature aggregation phase, which not only preliminarily adapts the source model to the target domain but also builds a feature distribution well-prepared for box prompt search. In the second phase, based on two feature distribution observations, we gradually expand the box prompt with the guidance of the target model feature and the SAM feature to handle the class-wise clustered target features and the class-wise dispersed 
target features, respectively. To remove the potentially enlarged false positive regions caused by the over-confident prediction of the target model, the refined pseudo-labels produced by SAM are further postprocessed based on connectivity analysis. 
Experiments on 3D and 2D datasets indicate that our approach yields superior performance compared to conventional methods. \revise{Code is available at \textcolor{blue}{\url{https://github.com/xmed-lab/DFG}}.}
\end{abstract}


\section{Introduction}
\label{sec:introduction}
Deep learning has exhibited excellent performance in medical image segmentation that plays an important role in clinical diagnosis and disease monitoring\cite{unet,wang2022medical}. However, deep neural networks are vulnerable to domain shifts caused by different scanning devices or imaging protocols. Data distribution discrepancy between the source and target domains normally leads to a dramatic performance degradation \revise{\cite{huang2025eval}},\cite{wang2019patch}, \revise{presenting major barriers to the clinical application of models.}

In recent years, Source-Free Domain Adaptation (SFDA) has been proposed to address the domain shift issue. SFDA has become a significant area of research \cite{shot,de,prabhu,ug,crs}, as it does not need to access the source data during domain adaptation, which well ensures the privacy of sensitive data. \revise{Moreover, it can significantly reduce the transmission burden associated with domain adaptation, and adapt public pre-trained models to downstream datasets.} SFDA aims to adapt a model pre-trained with source-domain images to align the target data distribution without requiring any annotation in the target domain. Recent SFDA methods for medical image segmentation generate pseudo-labels and conduct uncertainty-aware self-training \cite{dpl,ud4r,upl}, approximate source-style images \cite{fsm,fvp}, or make the source model more confident in its prediction by uncertainty reduction \cite{adami,protocontra}. However, all the existing SFDA strategies rely solely on source model and target data, which inevitably incurs errors under domain shift, thereby limiting adaptation performance. Thus, utilizing extra knowledge, which often refers to leveraging publicly available foundation models, is one promising solution to achieving higher performance.

Pre-trained on large segmentation datasets, Segment Anything Model (SAM) \cite{sam} is equipped with rich and generic knowledge and demonstrates impressive zero-shot segmentation performance given prompts such as bounding boxes or points. This motivates us to utilize SAM to assist domain adaptation for segmentation. \revise{However, as the prompts are by default interactively provided by humans, directly applying SAM in the target domain deviates from the goal of SFDA, which is intended for fully automatic adaptation for a segmentation task. Moreover, manually determining the box prompts requires expert knowledge and is labor-intensive.} 

\revise{In general, there are two ways of exploiting SAM in the SFDA scenario.} One is to finetune SAM (usually by injecting adapters), with the algorithms in, e.g., SAMed \cite{SAMed}, MA-SAM \cite{ma-sam} and Med-SA \cite{med-sa}. However, in SFDA where the ground truth is unavailable, the performance of finetuning SAM highly depends on the quality of pseudo-labels. Bad pseudo-labels cannot yield satisfactory results (see Section~\ref{ablation_finetune}). Another manner of leveraging SAM for SFDA is generating superior segmentation masks with the counterparts of SAM in the medical domain, e.g., MedSAM \cite{medsam} and SAM-Med3D \cite{sam-med3d}, to improve SFDA results, where the bottleneck is \revise{auto-generated} valid prompts. Among the prompting modes of the medical domain SAM, the bounding box prompt can clearly specify the segmentation target without ambiguity and yields better segmentation accuracy compared to the point prompt\cite{medsam,sam-med3d}. Thus, we aim to offer bounding box prompts to SAM, such as MedSAM\cite{medsam}, for enhancing domain adaptation. \revise{Existing auto-prompting methods require supervised fine-tuning of SAM, which falls outside the scope of the SFDA paradigm\cite{lin2024beyond,medsamu}. Naively using the predictions of the model resulting from existing SFDA methods to obtain box prompts for SAM shows limited performance gains (see the rows ``ProtoContra w/ MedSAM'' in Table~\ref{quanti_result_abdominal1},~\ref{quanti_result_abdominal2},~\ref{quanti_result_prostate}).
Therefore, a better method of automatically providing the bounding box prompt needs to be designed.} This paper focuses on using MedSAM, a renowned medical domain SAM with box prompt.

\begin{figure}[!t]
\centering
\includegraphics[width=\columnwidth]{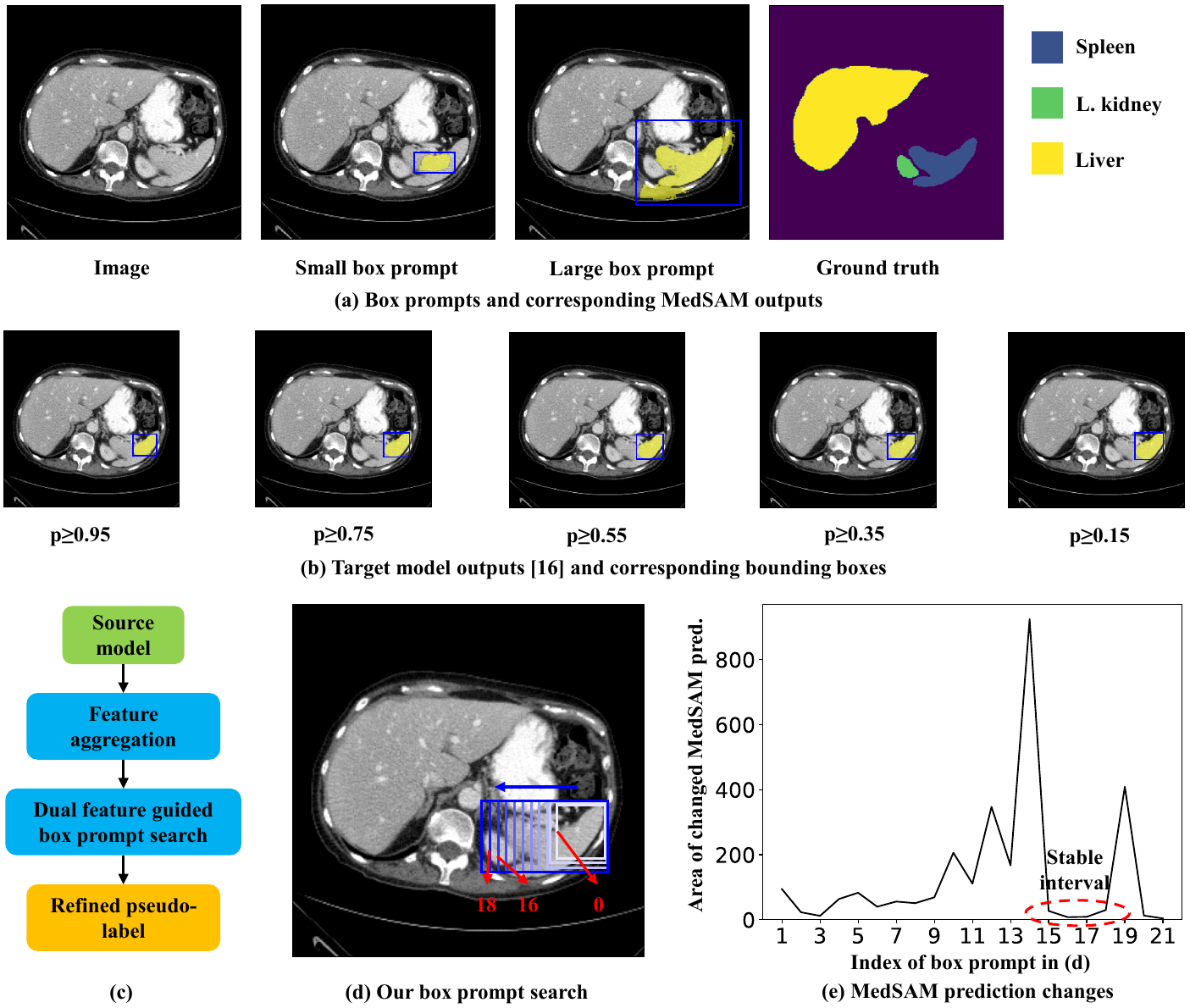}
\caption{Take spleen in a target domain image in MRI$\rightarrow$CT adaptation as an example. (a) MedSAM\cite{medsam} requires an accurate bounding box prompt. Neither a too-small nor a too-large bounding box leads to a decent segmentation result. (b) Segmentation results of \cite{protocontra} and the corresponding bounding boxes, produced by different output probability thresholds. Due to the domain gap and limited knowledge from source model and target unlabeled data, it is hard for existing SFDA methods to generate precise box prompts even if we vary the probability threshold. (c) After feature aggregation, our dual feature guided bounding box prompt search approach can find an accurate box prompt for MedSAM to yield refined pseudo-labels. \revise{(d) The searching procedure of our proposed box prompt search method. The red numbers are indices of the boxes, corresponding to the horizontal axis in (e). (e) The number of pixels of changed MedSAM predictions when the box prompt is switched from the last to the current. MedSAM prediction keeps stable when the box prompt fluctuates near the ground truth. We utilize this property to find the optimal box prompt.}} \label{intro}
\end{figure}

MedSAM requires the box prompt to be precise. Otherwise, its segmentation performance would severely degrade (see Fig.~\ref{intro} (a)). 
We find directly applying existing SFDA methods has difficulty finding a proper box prompt for a segmentation target (see Fig.~\ref{intro} (b)). Besides, we observe that the output of MedSAM would become relatively stable when the box prompt is perturbed by a few pixels from the ground truth (see Fig.~\ref{intro} (d)(e)). Hence, we propose to gradually expand the box prompt and choose the one in the stable interval.


In this paper, we propose a novel Dual Feature Guided (DFG) approach to search for the correct bounding box prompt to gain refined pseudo-labels. See Fig.~\ref{intro} (c) for a sketch of refining pseudo-labels, which are then postprocessed and used to train the target model. Our key idea is guiding the prompt search with two feature spaces based on two observations: (1) The source model is advantageous in that it shares the same label space with the target domain. On the other hand, the target domain features of the same class produced by the source model tend to be closely located despite the domain gap, which we refer to as feature affinity property. This property is partially preserved after the feature aggregation phase; (2) MedSAM does not have task-specific category semantics, but it has better feature affinity property than the target domain model.
Specifically, a feature aggregation phase prepares for the subsequent box prompt search by pushing features to their plausible class prototype to form clusters while partially maintaining the aforementioned feature affinity property. After the preparation, the box prompt is gradually expanded with propagation over the target model feature space and MedSAM feature space, to deal with the class-wise clustered target features and the class-wise dispersed target features, respectively. The propagation will terminate when a stable interval of MedSAM output is found. Fig.~\ref{intro} (d) shows an example of the box prompt search procedure. Finally, to eliminate the potentially enlarged false positive regions arising from the over-confident predictions of target model, the refined pseudo-labels generated by MedSAM are post-processed based on connectivity analysis. The main contributions of this paper are summarized as follows:
\begin{itemize}

\item We pioneeringly leverage SAM, e.g. MedSAM, for refining pseudo-labels in SFDA, taking advantage of the rich and generic knowledge of SAM.
\item We propose a Dual Feature Guided (DFG) auto-prompting approach to find the accurate bounding box prompt after a feature aggregation phase. DFG is based on the property of target model feature distribution under domain gap and the feature distribution characteristics of SAM. Besides, a connectivity-based post-processing makes the refined pseudo-labels robust to over-confident predictions of the target model.
\item Extensive experiments on 3D abdominal datasets and 2D prostate datasets show that our method greatly outperforms existing SFDA methods for medical image segmentation which only leverage the limited knowledge of source model and target data.

\end{itemize}

\section{Related Works}

\subsection{Source-Free Domain Adaptation for Medical Image Segmentation}
Existing SFDA solutions for medical image segmentation can be divided into four main categories: pseudo-labeling \cite{dpl,ud4r,cbmt,upl}, generating source-style images \cite{fsm,fvp}, uncertainty reduction \cite{adami,protocontra}, and batch normalization (BN) statistics adaptation \cite{bnmiccai}.
Pseudo-labeling methods generate pseudo-labels and then discard or correct erroneous pseudo-labels. For example, UPL\cite{upl} obtains pseudo-labels and identifies reliable ones by duplicating model heads with perturbations. U-D4R\cite{ud4r} performs uncertainty-weighted soft label correction by estimating the class-conditional label
error probability. Generating source-style images approaches aim to transfer a target domain image to source style while maintaining the image contents. For instance, FVP\cite{fvp} learns a visual prompt to add to a target image so that the frozen source model can perform well on their sum. Uncertainty reduction methods adapt the model by encouraging more confident predictions. For example, AdaMI\cite{adami} minimizes the entropy with a class-ratio regularization. For another instance, ProtoContra\cite{protocontra} minimizes the distance between target features and class prototypes with a bi-directional transport loss. For BN statistics adaptation methods, the parameters in BN layers are adapted to deal with the statistics discrepancy between different domains. Liu et al.\cite{bnmiccai} update low-order and high-order BN statistics with distinct training objectives. Notably, all of these methods only make use of the biased and limited knowledge of source model and unlabeled target data, thereby limiting their adaptation performance.

\vspace{-0.7em}
\subsection{Vision Foundation Models}
Vision foundation models have recently engaged a huge amount of attention in the computer vision community. Multimodal vision-language models, such as CLIP \cite{clip} and ALIGN \cite{align} learn from image-text pairs collected from the internet via contrastive learning. They have shown promising results across various vision or multimodal tasks. Later on, image segmentation foundation models, including SAM\cite{sam} and SegGPT\cite{seggpt}, are developed for general-purpose image segmentation. These large vision models have achieved awesome zero-shot generalization ability. Nowadays, the medical versions of SAM have also emerged for universal medical image segmentation. Among these models, MedSAM \cite{medsam} stands out because of its superior zero-shot segmentation ability using a bounding box prompt. Pre-trained with over one million images, MedSAM is equipped with astonishing segmentation capability across numerous modalities and domains. We pay particular attention to MedSAM in this work. Nonetheless, our method can be extended to other variants of SAM with box prompts.

\vspace{-0.7em}
\subsection{Leveraging Vision Foundation Models for Source-Free Domain Adaptation}
Vision foundation models are not source-biased and have better target domain generalizability. Consequently, a few works have tried integrating vision foundation models for SFDA. Co-learn\cite{rethink} integrates the pre-trained network, e.g. Swin Transformer into the target adaptation process, to leverage the more class-discriminative target representations. It distills useful knowledge in the pre-trained network through a co-learning algorithm to boost the pseudo-label quality. DALL-V\cite{dallv} exploits CLIP that contains the rich world prior robust to domain shift for source-free video domain adaptation. It distills the source/target domain adapted CLIP and original CLIP to a student network. DIFO\cite{sfda_clip} harnesses frozen CLIP, and devises a method that alternates between customizing CLIP through prompt learning and distilling the knowledge of customized CLIP to the target model.
Importantly, all of the mentioned methods aim at SFDA for natural image/video classification and thus utilize foundation models for classification which have different input formats and feature distribution property from those of the foundation models for segmentation. Therefore, these methods can not be easily extended to SFDA for medical image segmentation.

\begin{figure}[!t]
\centering
\includegraphics[width=0.9\columnwidth]{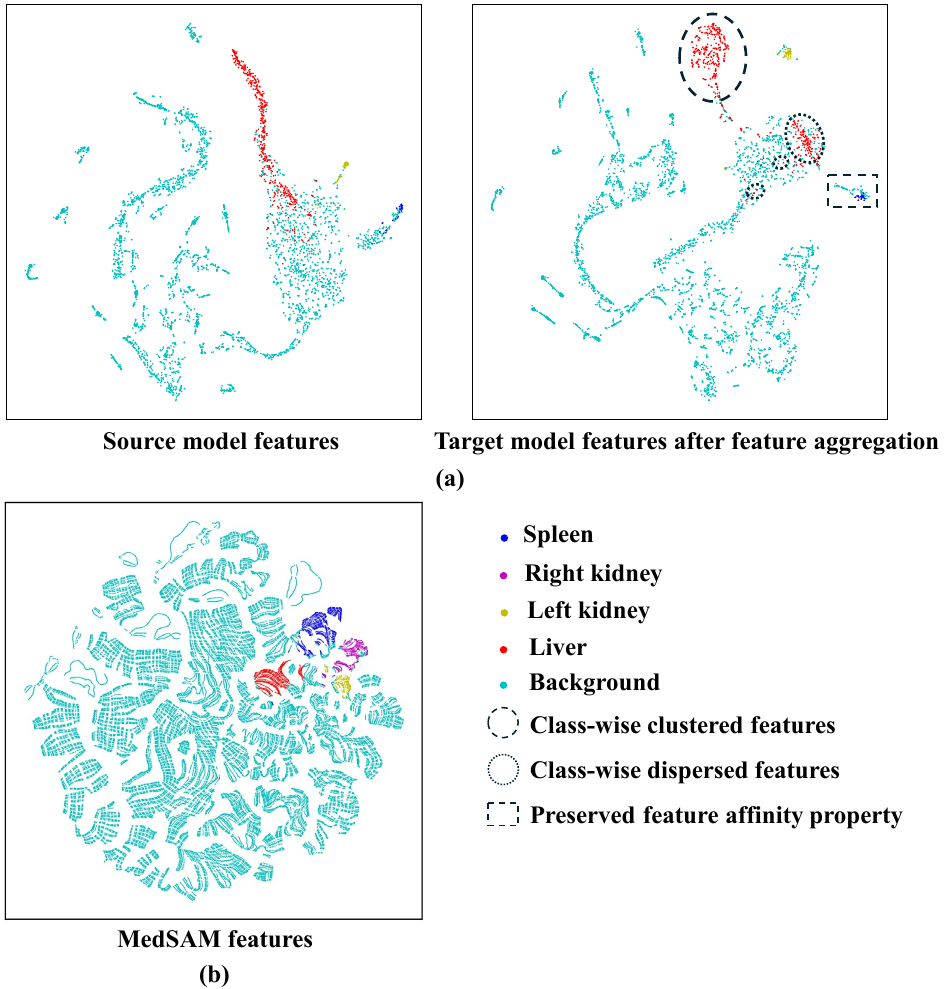}
\caption{T-SNE visualization of pixel features. (a) The features of a target domain image produced by the source model (left) and \revise{the target model after the feature aggregation phase (right)}. (b) The features of a target domain image produced by MedSAM.} \label{sec3}
\end{figure}

\section{Two Feature Distribution Properties}
\label{section3}
Our proposed DFG is based on the following two feature distribution properties that we observe in our experiments using t-SNE tool:

\begin{enumerate}[leftmargin=*]
\item The target domain features of the same class produced by the source model tend to be closely located despite the domain gap (feature affinity property). Moreover, the feature aggregation phase aggregates features of each class, while partially maintain the feature affinity property.
\item Though MedSAM lacks task-specific category semantics, it has better feature affinity than the target domain model.
\end{enumerate}

Fig.~\ref{sec3} (a) visualizes the features of a target domain image extracted before the last convolution layer of the source model, and the target model after the feature aggregation phase, respectively. The feature affinity property of source model is discernible. Such a property has been exploited in \cite{neighborhood} and implicitly utilized in the clustering-based pseudo-labeling in \cite{bmd,crs}. In this paper, we further aggregate such a feature distribution to prepare for the propagation over feature spaces during the box prompt search phase. It is noticeable from the right part of Fig.~\ref{sec3} (a) that some target features of each class are correctly pulled together (class-wise clustered features), while some other features of one class are wrongly pushed to the cluster of another class (class-wise dispersed features). The class-wise clustered features bring two benefits: (1) the propagation over the features of the same class can be easily carried out; (2) the gap between clusters sets an upper bound for the propagation in the feature space. Moreover, although some background features are pushed to the cluster of foreground features, the abovementioned feature affinity property still holds. Hence, when the propagation crosses the boundary between two classes within the same cluster, the MedSAM output will remain relatively stable and the propagation over the target model feature space will be terminated.

For the class-wise dispersed features, we tackle them with propagation over the MedSAM feature space. To verify the second feature distribution property, Fig.~\ref{sec3} (b) visualizes the features of a target domain image extracted at the output of the encoder of the MedSAM model. Benefiting from the large and diverse pre-training dataset, MedSAM has a strong generalizability on the target domain, as shown by the closely located features of the same class. Though MedSAM does not distinguish well between different foreground classes, such a feature distribution is sufficient for our bounding box prompt search as long as MedSAM features can differentiate between foreground and background pixels, considering the demand of box prompt search is to expand the box prompt in the direction of foreground instead of background.

\section{Methodology}
\begin{figure*}[!t]
\centering
\includegraphics[width=\textwidth]{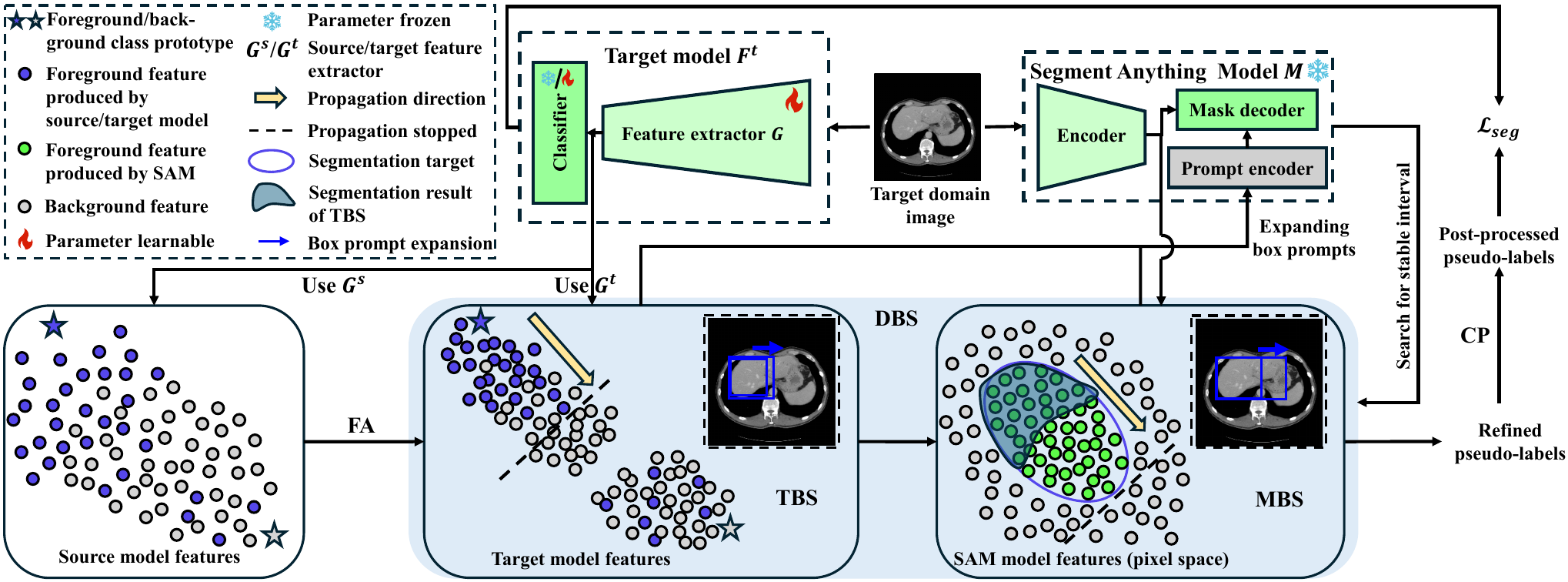}
\caption{\revise{Overview of our proposed Dual Feature Guided auto-prompting (DFG) framework. Feature aggregation (FA) forces target features to form clusters and prepares for box prompt search. Target model feature guided  box prompt search (TBS) and MedSAM feature guided box prompt search (MBS) find an accurate box prompt via propagation in the feature spaces to obtain refined pseudo-labels. Finally, the target model is trained with the connectivity-based post-processed (CP) pseudo-labels. The target model classifier is learnable during training with $\mathcal{L}_{seg}$ and frozen elsewhere.}} \label{method}
\end{figure*}

Fig.~\ref{method} illustrates our overall SFDA framework via dual feature guided auto-prompting.
In this section, we first describe the preliminaries of SFDA and MedSAM. Then, we introduce the feature aggregation phase that improves the target model performance and builds a feature distribution preparing for the box prompt search phase.
Next, we propose the dual feature guided bounding box prompt search strategy to find an accurate box prompt of each segmentation target for the refinement of pseudo-labels. Finally, we present the connectivity-based post-processing to eliminate the false-positive regions.

\vspace{-0.7em}
\subsection{Preliminaries}
Denote the source domain by $\mathcal{D}^s=(\mathcal{X}^s,\mathcal{Y}^s)$ and target domain by $\mathcal{D}^t=(\mathcal{X}^t,\mathcal{Y}^t)$.
In SFDA, a source model $F^s: \mathcal{X}^s\rightarrow\mathcal{Y}^s$ is trained with a source domain dataset $\{x^s_i,y^s_i\}_{i=1}^{n_s}$, where $(x^s_i,y^s_i) \in (\mathcal{X}^s,\mathcal{Y}^s)$. In this paper, we consider the medical image segmentation task, so $F^s$ is usually trained with a supervision loss like cross-entropy loss or Dice loss. $F^s$ consists of a feature extractor $G^s$ and a classifier $H^s$, i.e., $F^s(x)=H^s(G^s(x))$. $G^s$ maps each pixel $i \!\in\! \{1,\!\cdots\!,h\!\times\! w\}$ in an image to the feature $f_i\!\in\!\mathbb{R}^{d_F}$, and $H^s$ projects pixel feature into the semantic label space with $K$ classes. Here $h\times w$ is the image size and $d_F$ is the dimension of the extracted feature in model $F$. The source model is provided to the target client side. Also, an unlabeled dataset $\{x^t_i\}_{i=1}^{n_t}$ from the target domain $\mathcal{D}_t$ is given, where $x^t_i \in\mathcal{X}^t$. SFDA aims to learn a target model $F^t: \mathcal{X}^t\rightarrow\mathcal{Y}^t$ with only the source model $F^s$ and the target unannotated dataset $\{x^t_i\}_{i=1}^{n_t}$.

We use the MedSAM model $M$ to refine the pseudo-labels of the target dataset. $M$ takes as inputs a target image $x^t_i$ and a box prompt $P_{box}$ specifying the region of interest, and outputs a segmentation mask $y_M \in \{0,1\}^{h\times w}$. Besides, the encoder of MedSAM maps each pixel $i$ in the image to the feature $f^M_i\in\mathbb{R}^{d_M}$. $d_M$ is the feature dimension in model $M$.

\vspace{-0.7em}
\subsection{Feature Aggregation (FA)}\label{feature_aggre}
As shown in Section~\ref{section3}, due to the domain shift, the target features of the same class produced by the source model are dispersed and the target features of different classes  tend to be mixed, making it hard to propagate over the features of the same class. Feature aggregation seeks to force features to form clusters. As a result, part of the features of a foreground class are pulled together (class-wise clustered features), so that feature propagation can be carried out on these features to obtain part of the segmentation target. Besides, the gap between the clusters sets an upper bound for the propagation over the target model feature space.

To accomplish such an objective, we follow the conditional transport cost in \cite{zheng2021exploiting,protocontra}. The weights $c_k\!\in\! \mathbb{R}^{d_F}, k\!\in\!\{1,\!\cdots\!, K\}$ in the classifier $H^s$ of the source model can be regarded as the prototypes of each class \cite{shot}. \revise{We initialize the target model $F^t$ with the pre-trained source model parameters and freeze the classifier $H^s$ during feature aggregation.} Let $\langle\cdot,\cdot\rangle$ denote the cosine similarity. Consider a batch $\{x^t_b\}_{b=1}^{B}$ of $B$ images, then the cosine distance between the target model feature of the $i$-th pixel in the $b$-th image $f^t_{b,i}$ and $c_k$ is $d(c_k,f^t_{b,i})=1-\langle c_k,f^t_{b,i}\rangle$.
\revise{The probability of classifying $f^t_{b,i}$ as class $k$ and the probability of classifying $c_k$ as the $i$-th pixel in the $b$-th image are first computed as}
\revise{
\begin{equation}\label{feature_aggre_prob}
\begin{split}
p(y^t_{b,i}=k|f^t_{b,i})&=\frac{\text{exp}(c^T_kf^t_{b,i}/\kappa)}{\sum_{k'\!=\!1}^{K}\!\text{exp}(c^T_{k'}\!f^t_{b,i}/\!\kappa)},\\
p(y_{c_k}=(b,i)|c_k)&=\frac{\text{exp}(c^T_kf^t_{b,i}/\kappa)}{\sum_{b'\!=\!1}^{B}\!\sum_{i'\!=\!1}^{h\!\times\! w}\!\text{exp}(c^T_{k}\!f^t_{b',i'}/\!\kappa)},
\end{split}
\end{equation}
}
where $\kappa$ is the temperature constant. Then the feature aggregation loss is defined as
\revise{
\begin{equation}\label{feature_aggre_loss}
\begin{split}
\mathcal{L}_{fa}\!&=\!\frac{1}{B\!\times\! h\!\times\! w}\!\sum\limits_{b=1}^{B}\!\sum\limits_{i=1}^{h\times w}\!\sum\limits_{k=1}^{K}\!d(c_k,\!f^t_{b,i})p(y^t_{b,i}=k|f^t_{b,i})\\
&+\!\frac{1}{K}\!\sum\limits_{k=1}^{K}\!\sum\limits_{b=1}^{B}\!\sum\limits_{i=1}^{h\times w}\!d(c_k,\!f^t_{b,i})p(y_{c_k}=(b,i)|c_k).
\end{split}
\end{equation}
}
The first term in Eq.~\ref{feature_aggre_loss} is the average distance from each target feature to the class prototypes. As a result, each target feature will be pulled to a plausible class prototype. This will not only align the target features with the frozen source classifier, but also aggregate target features of each class. In the meantime, the feature affinity property of the source model is supposed to be partially preserved, as the loss is only to pull features to prototypes without any other constraints. On the other hand, to avoid trivial solutions where all the features are pulled to the same class prototype, the second term in Eq.~\ref{feature_aggre_loss} which measures the distance from each class prototype to the target features is added, ensuring each class prototype is assigned to some target features. After the feature aggregation phase, a target feature distribution with the first property stated in Section~\ref{section3} is formed, well preparing for the box prompt search phase.

\vspace{-0.7em}
\subsection{Dual Feature Guided Box Prompt Search (DBS)}
Based on the two feature distribution properties presented in Section~\ref{section3}, we propose to perform the box prompt search with the propagation on the target model features and the MedSAM features, to deal with the class-wise clustered features and the class-wise dispersed features, respectively.

\subsubsection{Target model feature guided box prompt search (TBS)}
The purpose of TBS is to find a bounding box prompt that covers all the pixels corresponding to the class-wise clustered features so that these pixels are partitioned by MedSAM. Based on the first feature distribution property observed in Section~\ref{section3} and the fact that MedSAM output becomes relatively stable when the box prompt varies around an organ boundary, we expand the box prompt via propagating over the target model feature space and terminate at the boundary between class-wise clustered features and background features.

For a 2D slice, the most confident pixels $\mathcal{I}_0$ of a foreground class $k$ are chosen as the inception of the box prompt search. $\mathcal{I}_0$ is obtained by simply thresholding the target model probabilities $p_{i,k}$, which denotes the model output probability of the $k$-th class at the $i$-the pixel. To cope with the class-imbalanced domain gap problem \cite{ud4r,bmd} in SFDA, an adaptive threshold is adopted to pick the pixels that have a probability difference with the maximum value less than $p_\Delta$, as in
\begin{equation}\label{i0}
\mathcal{I}_0=\{i\vert p_{i,k}>\max\limits_i(p_{i,k})-p_\Delta\}.
\end{equation}
The set of pixels $\mathcal{I}_j$ is gradually enlarged with the guidance of the target model features to ensure that the box prompt expands in the correct direction. Considering an organ/tissue in a medical image is usually consecutive, at each iteration, we look at the spatially adjacent pixels of the current $\mathcal{I}_j$, and those whose features are close to the features of pixels in $\mathcal{I}_j$ are added. Namely,
\begin{equation}\label{propagate1}
\mathcal{I}_{j+1}=\{i\vert \langle f^t_i,f^t_{i'}\rangle>\tau_f, d_E(i,i')<r,\forall i'\in\mathcal{I}_{j}\}.
\end{equation}
Here $f^t_i$ is the target model feature of the $i$-th pixel, $\tau_f$ is the threshold for feature similarity, $d_E(\cdot,\cdot)$ is the Euclidean distance, and $r$ is the parameter controlling how many pixels to propagate in one step. Since the feature aggregation phase has made the features of each class gather, \revise{$\tau_f$ is set to a constant of high value, which is empirically chosen as 0.99 across all the datasets in our experiments}.

As $\mathcal{I}_{j}$ incorporates more pixels, the corresponding target features are expanding over the target feature space. Accordingly, the bounding box prompt $P_{box,j}$ for the $j$-th iteration is also expanded. $P_{box,j}$ is set to cover $\mathcal{I}_{j}$ and has a margin of $m$ pixels to accommodate the case where there are a few pixels not aggregated to the class prototype (the case that there are not lots of class-wise dispersed features). A MedSAM prediction is calculated for each bounding box prompt, i.e., $y_{M,j}=M(x^t,P_{box,j})$. Then we compare the predictions in different iterations and look for the stable interval via
\begin{equation}\label{delta_M}
\Delta_M(j,j')=\sum\limits_i\mathbbm{1}(y_{M,j,i}\neq y_{M,j',i}),
\end{equation}
where $\mathbbm{1}$ is the indicator function, and $y_{M,j,i}$ is the prediction of MedSAM in the $j$-th iteration at the $i$-th pixel. $\Delta_M$ represents the total number of pixels where the MedSAM predictions in the two iterations differ. When the propagation over target model feature space crosses the boundary between the foreground features and the background (or the other foreground classes) features, the MedSAM output will become relatively stable, as indicated by Fig.~\ref{intro}(e). If $\Delta_M$ is less than a threshold $\tau_\Delta$, $j\rightarrow j'$ is regarded as a stable interval and the propagation is stopped. In practice, we priorly search for a long stable interval to avoid false stability. Hence, we search for a low $\Delta_M$ with $j'-j$ equal to 3, 2, 1 in succession. 

\subsubsection{MedSAM feature guided box prompt search (MBS)}
Due to the existence of class-wise dispersed features of the target model, the propagation over the target model features might not be able to cover all of the features of the class $k$. As a consequence, only partial pixels of class $k$ are recognized by MedSAM. To tackle the class-wise dispersed features of the target model and get the full segmentation of class $k$, we take advantage of the second property in Section~\ref{section3} to propagate over the MedSAM feature space.

Let $\mathcal{I}'_0$ denote the set of pixels partitioned by MedSAM at the end of the TBS. According to the above analysis, $\mathcal{I}'_0$ is assumed to contain part of the pixels of class $k$. Therefore, the prototype $c_M$ and feature divergence $\text{Div}_M$ of the class in the MedSAM feature space can be estimated as
\begin{equation}
c_M=\mathop{avg}\limits_{i\in\mathcal{I}'_0}(f^M_i/\Vert f^M_i\Vert),\
\text{Div}_M=\mathop{avg}\limits_{i\in\mathcal{I}'_0}d(f^M_i,c_M),
\end{equation}
where $d(\cdot,\cdot)$ is cosine distance as defined in Section~\ref{feature_aggre}, and $\mathop{avg}$ is the averaging operation.

Then $\mathcal{I}'_j$ is gradually expanded with the guidance of the MedSAM features. At each iteration, we look at the spatially adjacent pixels of $\mathcal{I}'_j$, and those whose features are close to
the prototype are added to $\mathcal{I}'_j$, as in
\begin{equation}\label{propagate2}
\begin{gathered}
\mathcal{I}'_{j+1}=\{i\vert  d(f^M_i,c_M)\!<\!\tau_d, 
d_E(i,i')\!<\!r,\forall i'\in\mathcal{I}'_{j}\}\cup\mathcal{I}'_{j},\\
\tau_d=\min(\tau_{Div}\cdot\text{Div}_M,\tau_{max}),
\end{gathered}
\end{equation}
\revise{where $f^M_i$ is the MedSAM feature of the $i$-th pixel, $d_E(\cdot,\cdot)$ is the Euclidean distance, and $r$ controls propagation step size.} $\tau_d$ is the distance threshold in propagation, and it is determined by a coefficient $\tau_{Div}$ and the feature divergence. $\tau_{Div}$ is set as 2.5 in all of our experiments, given that $\text{Div}_M$ is the average distance to the prototype in $\mathcal{I}'_0$. To prevent the case that $\text{Div}_M$ is so large that background pixels are also added, we set an upper bound $\tau_{max}$ for the distance threshold.
It is worth noting that the objective of bounding box search is to make the bounding box expand in the direction of foreground instead of background. Therefore, we use the feature divergence-based threshold to distinguish between foreground and background, rather than compare the distances to different class prototypes as in the typical classification task.

The remaining steps are similar to those in TBS. $P'_{box,j}$ is set to cover $\mathcal{I}'_{j}$ and the corresponding MedSAM output is calculated. Then the stable interval is found by seeking a low $\Delta_M$ in Eq.~\ref{delta_M} in the same manner as in the last subsubsection.

When the box prompt cannot further expand, which might be caused by an inappropriate distance threshold computed in Eq.~\ref{propagate2}, we artificially expand the four edges of the bounding box prompt by $\frac{r}{2}$ pixels for a couple of iterations and try to find a stable interval. If the stable interval is still not observed, the MedSAM segmentation mask in the last iteration before the artificial expansion of the bounding box prompt is deemed to be the final segmentation result of class $k$.

\vspace{-0.7em}
\subsection{Connectivity-based Post-processing (CP)}\label{cp}
The ultimate segmentation masks of MedSAM for each foreground class constitute the refined pseudo-labels. Nonetheless, the target model could be over-confident in its predictions. As a result, a small false positive region of a class $k$ predicted by the target model might cause a much larger area, e.g., a whole organ belonging to the background, to be treated as class $k$ after the DBS.

To address the issue of potentially increased false positive area, we post-process the refined pseudo-labels by only keeping the largest connected component for each foreground class, since an organ in a medical image ought to be consecutive. Finally, the target model $F^t$ is trained under the supervision of the post-processed pseudo-labels $\widetilde{y}^t$. The loss function is the Dice loss, as in
\begin{equation}
\mathcal{L}_{seg}=\mathcal{L}_{dice}(F^t(x^t),\widetilde{y}^t).
\end{equation}

\section{Experiments}
\subsection{Datasets and Implementation}
We verify our proposed DFG approach for SFDA of medical image segmentation using \revise{three} abdominal datasets and one prostate dataset as the target domain. For a fair experimental result, the target domain datasets should not be employed as the training sets of MedSAM. MedSAM has harnessed an exhaustive list of public datasets across a wide spectrum of tasks. To the best of our knowledge, these \revise{four} datasets are the only ones which are not used in the training of MedSAM \revise{and simultaneously have paired datasets with shared classes and noticeable domain discrepancies}.

\subsubsection{The abdominal datasets}
Our proposed method is first evaluated on a cross-modality abdominal multi-organ segmentation task, where the Beyond the Cranial Vault (BTCV) challenge dataset\cite{miccai2015}, the CHAOS\cite{chaos} dataset, \revise{and the CURVAS challenge dataset\cite{curvas}} are involved. We utilize the 30 CT volumes from the BTCV dataset, the 20 T2-SPIR MRI volumes from the CHAOS dataset, \revise{and the 25 CT volumes from the training and validation cohorts of CURVAS dataset}. Experiments are performed in both MRI to CT direction and CT to MRI direction. \revise{The domain adaptation between BTCV and CHAOS involve four abdominal organs: spleen, right kidney, left kidney, and liver, whereas the domain adaptation between CHAOS and CURVAS are conducted on their two shared foreground classes: kidney and liver}. Following \cite{protocontra,fvp}, \revise{we randomly divide BTCV and CHAOS into training and test
sets with a ratio of 4:1, and use the training/validation subsets of CURVAS as training/test sets}. The axial slices that do not contain any foreground class are discarded \cite{protocontra,bian2021domain}. The pixel values are clipped to $[-125, 275]$ and $[0, 1200]$ for the \revise{CT and MRI} datasets, respectively, followed by min-max normalization which normalizes the values to $[0, 1]$. All the volumes are resampled into the size of $256\times256$ in axial plane. Each volume is split into axial view slices \cite{protocontra,chen2020unsupervised}. The BTCV \revise{and CURVAS} datasets are not among the curated datasets of MedSAM\footnote{\url{https://github.com/bowang-lab/MedSAM/blob/main/assets/MedSAM_supp.pdf}}. The CHAOS is used as ``external validation set''\footnote{\revise{This term is taken from MedSAM paper, meaning hold-out datasets to test MedSAM’s generalization ability on unseen datasets or segmentation targets}.} of MedSAM, \revise{which is not used during MedSAM training or hyperparameter tuning~\cite{medsam}}.


\begin{table*}[t]
\centering
\caption{Quantitative Comparison of the Segmentation Results with State-of-the-arts on the Abdominal Datasets. ``SAM" means using SAM during adaptation. The $\uparrow$ Sign Indicates a Higher Value is Better, and the $\downarrow$ Sign Indicates a Lower Value is Better. The Bold Font Highlights the Best Results.}\label{quanti_result_abdominal1}
\begin{tabular}{cc|ccccc|ccccc}

\toprule[1.5pt]

\multicolumn{12}{c}{\textbf{Source: CHAOS; Target: BTCV} (\textbf{MRI}$\rightarrow$\textbf{CT})}\\\hline
\multirow{2}{*}{Methods} & \multirow{2}{*}{\textbf{SAM}} & \multicolumn{5}{c|}{\textbf{Dice}(\%)$\uparrow$} & \multicolumn{5}{c}{\textbf{ASSD}$\downarrow$}\\\cline{3-12}
&& Liver & R. Kidney & L. Kidney & Spleen & Average & Liver & R. Kidney & L. Kidney & Spleen & Average \\
\hline

Source model&-& 32.6 & 33.3 & 51.9 & 41.1 & 39.7 & 6.73 & 14.75 & 9.27 & 9.90 & 10.16\\
Target supervised &-& 93.3 & 93.4 & 92.4 & 88.1 & 91.8 & 3.10 & 0.52 & 0.41 & 1.37 & 1.35\\\hline
DPL\cite{dpl} & \ding{55} & 39.0 & 35.3 & 58.6 & 42.4 & 43.9 & 6.36 & 13.09 & 7.94 & 10.10 & 9.38\\
AdaMI\cite{adami} & \ding{55} & 72.8 & 57.0 & 63.5 & 47.1 & 60.1 & 5.51 & 11.54 & 10.7 & 10.78 & 9.64\\
UPL\cite{upl} & \ding{55} & 78.1 & 61.0 & 68.2 & 38.9 & 61.5 & 4.21 & 6.80 & 6.95 & 9.31 & 6.82\\
FVP\cite{fvp} & \ding{55} & 87.8 & 64.7 & 73.2 & 68.3 & 73.5 & 3.63 & 2.58 & 3.10 & 2.34 & 2.91\\
ProtoContra\cite{protocontra} & \ding{55} & 85.4 & 75.5 & 68.0 & 68.9 & 74.4 & 3.22 & 5.91 & 6.17 & \textbf{4.18} & 4.87\\
ProtoContra w/ MedSAM & \checkmark & 85.1 & 78.0 & 71.5 & 66.6 & 75.3 & 3.38 & 4.33 & 5.98 & 5.09 & 4.70\\
DFG (ours) & \checkmark & \textbf{90.9} & \textbf{88.9} & \textbf{82.7} & \textbf{77.3} & \textbf{84.9} & \textbf{1.29} & \textbf{0.66} & \textbf{1.74} & 4.28 & \textbf{1.99}\\
\midrule[1.5pt]

\multicolumn{12}{c}{\textbf{Source: BTCV; Target: CHAOS} (\textbf{CT}$\rightarrow$\textbf{MRI})}\\\hline
\multirow{2}{*}{Methods} &\multirow{2}{*}{\textbf{SAM}}& \multicolumn{5}{c|}{\textbf{Dice}(\%)$\uparrow$} & \multicolumn{5}{c}{\textbf{ASSD}$\downarrow$}\\\cline{3-12}
&& Liver & R. Kidney & L. Kidney & Spleen & Average & Liver & R. Kidney & L. Kidney & Spleen & Average \\
\hline

Source model&-& 53.1 & 46.9 & 40.9 & 20.4 & 40.3 & 3.84 & 1.66 & 2.49 & 4.23 & 3.06\\
Target supervised &-& 92.4 & 93.2 & 91.1 & 76.3 & 88.2 & 0.93 & 0.61 & 0.35 & 1.44 & 0.83\\\hline
DPL\cite{dpl} & \ding{55} & 63.1 & 65.3 & 57.3 & 38.3 & 56.0 & 2.86 & 1.14 & 1.40 & 2.61 & 2.00\\
AdaMI\cite{adami} & \ding{55} & 54.8 & 77.2 & 64.4 & 47.6 & 61.0 & 4.90 & 5.75 & 4.02 & 5.87 & 5.14\\
UPL\cite{upl} & \ding{55} & 56.9 & 55.6 & 59.9 & 55.3 & 56.9 & 3.86 & 1.47& 1.58 & 4.77 & 2.92\\
FVP\cite{fvp} & \ding{55} & 64.8 & 87.6 & 80.3 & 60.5 & 73.3 & 4.48 & 2.10 & 1.54 & 6.15 & 3.57\\
ProtoContra\cite{protocontra} & \ding{55} & 71.6 & 83.8 & \textbf{81.6} & 84.9 & 80.5 & 2.89 & 3.36 & 4.73 & 1.09 & 3.02\\
ProtoContra w/ MedSAM & \checkmark & 78.4 & 72.0 & 65.0 & 84.6 & 75.0 & 2.30 & 4.46 & 8.40 & 1.07 & 4.06\\
DFG (ours) & \checkmark & \textbf{83.6} & \textbf{92.5} & 79.6 & \textbf{85.1} & \textbf{85.2} & \textbf{1.82} & \textbf{0.30} & \textbf{0.87} & \textbf{1.02} & \textbf{1.00}\\

\bottomrule[1.5pt]

\end{tabular}
\end{table*}

\subsubsection{The prostate datasets}
We employ the NCI-ISBI dataset \cite{nci-isbi,liu2020ms} and the QUBIQ dataset \cite{qubiq} for prostate segmentation. For the NCI-ISBI dataset, 30 MRI volumes from the Radboud University Nijmegen Medical Centre are used, which are randomly separated into training/test sets with a ratio of 4:1. Each volume is split as slices in axial plane. The QUBIQ dataset contains 55 MRI cases for prostate segmentation. For each case, a representative slice has been selected by the challenge organizer, so QUBIQ can be viewed as a 2D dataset. We adopt its original training/test split. The pixel values of these two datasets are clipped to the range between the 0.5th and 99.5th percentiles before rescaling them to the range of $[0, 1]$\cite{medsam}. Each slice is resized to $256\times256$. QUBIQ is used as external validation set of MedSAM, \revise{which is not used during the training or hyperparameter tuning of MedSAM}, whereas NCI-ISBI is part of the training set of MedSAM. Thus, we conduct adaptation experiments only in the “NCI-ISBI to QUBIQ” direction.

\begin{table}[t]
\centering
\caption{\revise{Quantitative Comparison of the Segmentation Results with State-of-the-arts on the Abdominal Datasets (CHAOS to CURVAS Adaptation, {MRI}$\rightarrow${CT}).} }\label{quanti_result_abdominal2}
\begin{tabular}{C{2cm}@{\hspace{2pt}}C{0.65cm}@{\hspace{2pt}}|@{\hspace{2pt}}C{0.8cm}@{\hspace{2pt}}C{0.9cm}@{\hspace{2pt}}C{0.7cm}|@{\hspace{2pt}}C{0.8cm}@{\hspace{2pt}}C{0.9cm}@{\hspace{2pt}}C{0.7cm}@{\hspace{2pt}}}

\toprule[1.5pt]

\multirow{2}{*}{Methods} & \multirow{2}{*}{\textbf{SAM}} & \multicolumn{3}{C{2.4cm}|@{\hspace{2pt}}}{\textbf{Dice}(\%)$\uparrow$} & \multicolumn{3}{c}{\textbf{ASSD}$\downarrow$}\\\cline{3-8}
&& Liver & Kidney & Avg. & Liver & Kidney & Avg. \\
\hline

Source model&-& 2.1 & 21.0 & 11.6 & 27.05 & 8.34 & 17.69\\
Target supervised &-& 96.0 & 94.0 & 95.0 & 0.56 & 2.01 & 1.28 \\\hline
DPL\cite{dpl} & \ding{55} & 3.1 & 28.8 & 15.9 & 30.76 & 8.35 & 19.56 \\
AdaMI\cite{adami} & \ding{55} & 75.7 & 45.2 & 60.5 & 3.78 & 9.51 & 6.65 \\
UPL\cite{upl} & \ding{55} & 87.8 & 50.9 & 69.3 & 2.52 & 9.70 & 6.11 \\
ProtoContra\cite{protocontra} & \ding{55} & 88.0 & 67.2 & 77.6 & 2.39 & 7.20 & 4.79 \\
ProtoContra w/ MedSAM & \checkmark & 92.5 & 71.4 & 81.9 & 1.72 & 6.82 & 4.27 \\
DFG (ours) & \checkmark & \textbf{94.0} & \textbf{77.4} & \textbf{85.7} & \textbf{1.02} & \textbf{3.53} & \textbf{2.27} \\

\bottomrule[1.5pt]

\end{tabular}
\end{table}

\begin{table}[t]
\centering
\caption{Quantitative Comparison of the Segmentation Results with State-of-the-arts on the Prostate Datasets (NCI-ISBI to QUBIQ adaptation).}\label{quanti_result_prostate}
\begin{tabular}{cc|c|c}

\toprule[1.5pt]

Methods & \textbf{SAM} & \textbf{Dice}(\%)$\uparrow$ & \textbf{ASSD}$\downarrow$\\\hline

Source model &-& 66.9 & 10.64\\
Target supervised &-& 94.4 & 1.86\\\hline
DPL\cite{dpl} & \ding{55} & 75.5 & 8.33\\
AdaMI\cite{adami} & \ding{55} & 81.0 & 8.46\\
UPL\cite{upl} & \ding{55} & 73.6 & 16.55\\
ProtoContra\cite{protocontra} & \ding{55} & 86.6 & 5.58\\
ProtoContra w/ MedSAM & \checkmark & 83.0 & 5.11\\
DFG (ours) & \checkmark & \textbf{93.3} & \textbf{2.18}\\

\bottomrule[1.5pt]

\end{tabular}
\end{table}

\begin{figure*}[!t]
\centering
\includegraphics[width=0.72\textwidth]{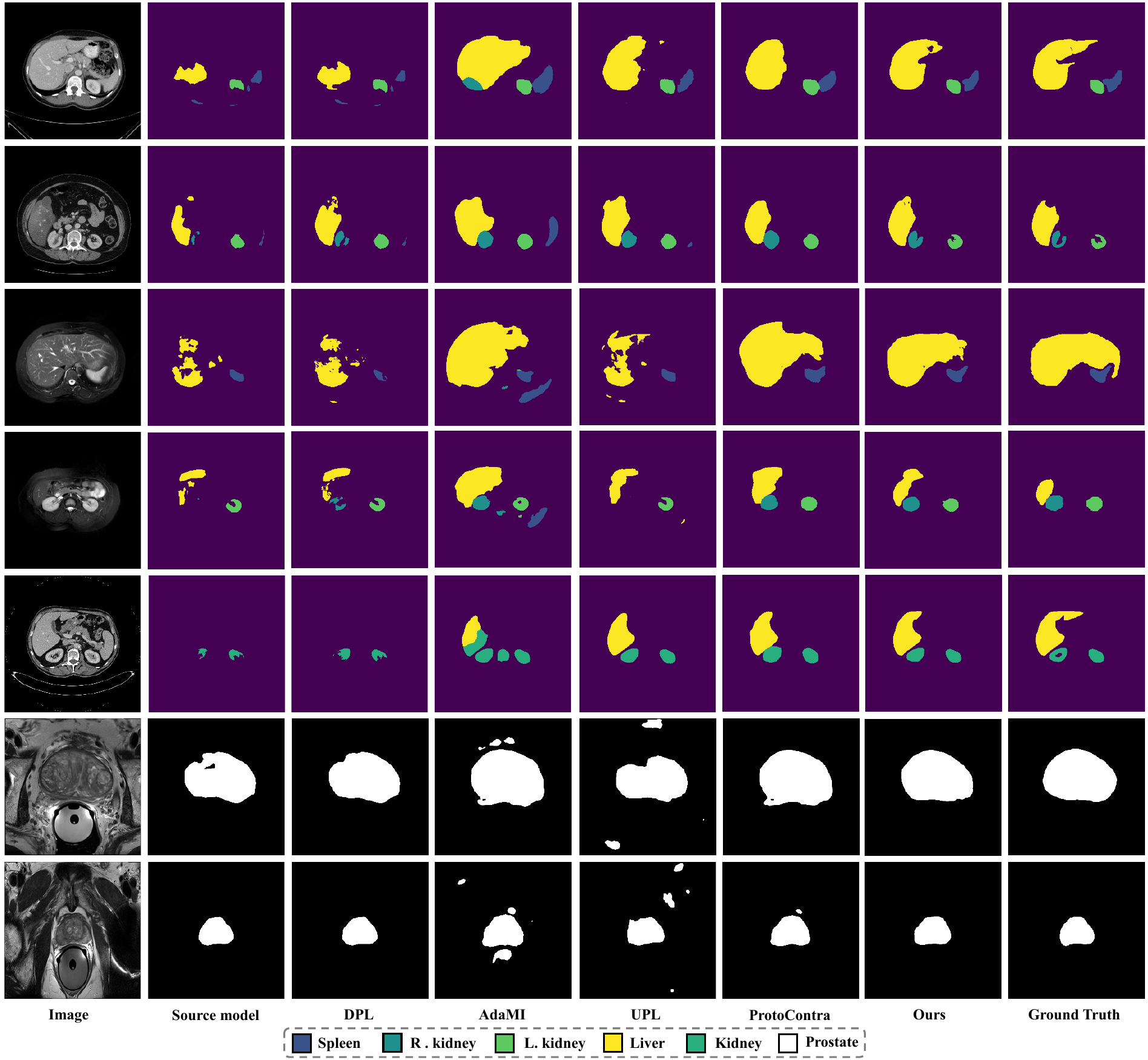}
\caption{\revise{Qualitative comparison of different SFDA methods. The top two rows are segmentation results of BTCV images in CHAOS$\rightarrow$BTCV. The 3rd and 4th rows are segmentation results of CHAOS images in BTCV$\rightarrow$CHAOS. The 5th row is segmentation results of Curvas image in CHAOS$\rightarrow$Curvas. The bottom two rows are segmentation results of prostate images.}} \label{quali}
\end{figure*}

\subsubsection{Implementation details}
We adopt the U-Net\cite{unet} as the segmentation network. During domain adaptation, the model is optimized using the Adam optimizer with a learning rate of $1\times10^{-4}$ and a weight decay of $5\times10^{-4}$. The model is trained for 5 epochs during the feature aggregation phase and 100 epochs in the training with the post-processed pseudo-label, with a batch size of 16. The temperature $\kappa$ is 1 for the abdominal datasets and 10 for the prostate dataset. The feature similarity threshold $\tau_f$ in Eq.~\ref{propagate1} and the distance threshold coefficient $\tau_{Div}$ in Eq.~\ref{propagate2} are set to 0.99 and 2.5 respectively for all the datasets. The $p_\Delta$ in Eq.~\ref{i0} and the $r$ controlling propagation step size are empirically set as $0.5\%$ and 4. The $\tau_{max}$ in Eq.~\ref{propagate2} is 0.35 for the abdominal datasets and 0.3 for the prostate dataset. The threshold $\tau_\Delta$ for $\Delta_M(j,j')$ in Eq.~\ref{delta_M} is set to 15, 30, 45 for $j'-j$ being 1, 2, 3, respectively. Referring to \cite{protocontra}, Data augmentations including random cropping, rotation, and brightness adjustment are applied. All the experiments are implemented with PyTorch on a single RTX 3090 GPU.

\subsubsection{Evaluation metrics}
For quantitative evaluation of the segmentation results, two common metrics, Dice coefficient (Dice) and
average symmetric surface distance (ASSD) are adopted. 3D Dice score is used for the 3D abdominal datasets and 2D Dice score is used for the 2D prostate dataset. Follow \cite{protocontra,fvp}, we calculate ASSD values with unit of pixel.

\vspace{-0.7em}
\subsection{Comparison with State-of-the-Art Methods}
To validate the effectiveness of our proposed DFG, we compared it with five recent state-of-the-art (SOTA) SFDA for medical image segmentation methods: 
DPL \cite{dpl}, AdaMI \cite{adami}, UPL \cite{upl}, FVP \cite{fvp}, and ProtoContra \cite{protocontra}.
DPL \cite{dpl} denoised pseudo-labels via pixel uncertainty estimation and prototype estimation. AdaMI \cite{adami} minimizes the entropy of model predictions with a class-ratio regularizer. UPL \cite{upl} obtain pseudo labels and their uncertainty via diversified predictions. FVP \cite{fvp} learns a visual prompt parameterized with some low-frequency parameters in the frequency space. ProtoContra \cite{protocontra} aligns target features with class prototypes and uses contrastive learning to train the pixels with unreliable predictions.
\revise{For a fair comparison, we reimplement DPL, AdaMI, UPL, and ProtoContra based on their papers and released codes. We utilize the same segmentation network, data augmentations, and type of optimizers. The batch sizes are tuned to choose the optimal values, and all the other hyperparameters are unchanged.}  For FVP, we directly reference the results from their paper as the codes are not released.

In addition, we also compare our approach with some naive methods. ``Source model" denotes directly applying the pre-trained source model to the target data without any adaptation. ``Target supervised" means training a target domain model in a fully supervised manner, which acts as an upper bound. ``w/ MedSAM" stands for directly exploiting a SOTA method's predictions to obtain the bounding boxes which are then used as prompts of MedSAM for refinement of the predictions. ``w/ MedSAM" serves as a baseline for using MedSAM in SFDA.

\begin{table*}[t]
\centering
\caption{Ablation Study on Different Components of Our Approach. }\label{ablation_component}
\begin{tabular}{cccc|ccccc|ccccc}

\toprule[1.5pt]
\multicolumn{4}{c|}{Components}&\multicolumn{5}{c|}{\textbf{Dice}(\%)$\uparrow$}
& \multicolumn{5}{c}{\textbf{ASSD}$\downarrow$}\\\hline

\multirow{2}{*}{FA} & \multicolumn{2}{|c|}{DBS} &\multirow{2}{*}{CP} & \multirow{2}{*}{Liver} & \multirow{2}{*}{R. Kidney} & \multirow{2}{*}{L. Kidney} & \multirow{2}{*}{Spleen} & \multirow{2}{*}{Average} & \multirow{2}{*}{Liver} & \multirow{2}{*}{R. Kidney} & \multirow{2}{*}{L. Kidney} & \multirow{2}{*}{Spleen} & \multirow{2}{*}{Average}\\\cline{2-3}

\multicolumn{1}{c|}{}& \multicolumn{1}{c|}{TBS} & \multicolumn{1}{c|}{MBS} & &  &  &  &  &  &  &  &  &  &  \\
\hline

\ding{55}&\ding{55}&\ding{55}&\ding{55}& 32.6 & 33.3 & 51.9 & 41.1 & 39.7 & 6.73 & 14.75 & 9.27 & 9.90 & 10.16\\
\checkmark &\ding{55}&\ding{55}&\ding{55} & 84.2 & 73.1 & 73.1 & 62.5 & 73.2 & 4.35 & 6.73 & 5.70 & 5.93 & 5.68\\
\checkmark & \checkmark &\ding{55}&\ding{55} & 85.3 & 81.7 & 80.5 & 74.9 & 80.6 & 3.03 & 6.48 & 3.71 & 5.54 & 4.69\\
\checkmark & \checkmark & \checkmark &\ding{55} & \textbf{91.1} & 83.0 & 81.9 & 75.7 & 82.9 & \textbf{1.21} & 6.53 & 3.72 & 4.90 & 4.09\\
\checkmark & \checkmark &\ding{55}& \checkmark & 84.9 & 86.5 & \textbf{84.0} & 74.9 & 82.6 & 3.10 & 0.74 & \textbf{1.10} & 4.66 & 2.40\\

\checkmark & \checkmark & \checkmark & \checkmark & 90.9 & \textbf{88.9} & 82.7 & \textbf{77.3} & \textbf{84.9} & 1.29 & \textbf{0.66} & 1.74 & \textbf{4.28} & \textbf{1.99}\\
\bottomrule[1.5pt]

\end{tabular}
\end{table*}

\begin{figure*}[t]
\centering
\includegraphics[width=\textwidth]{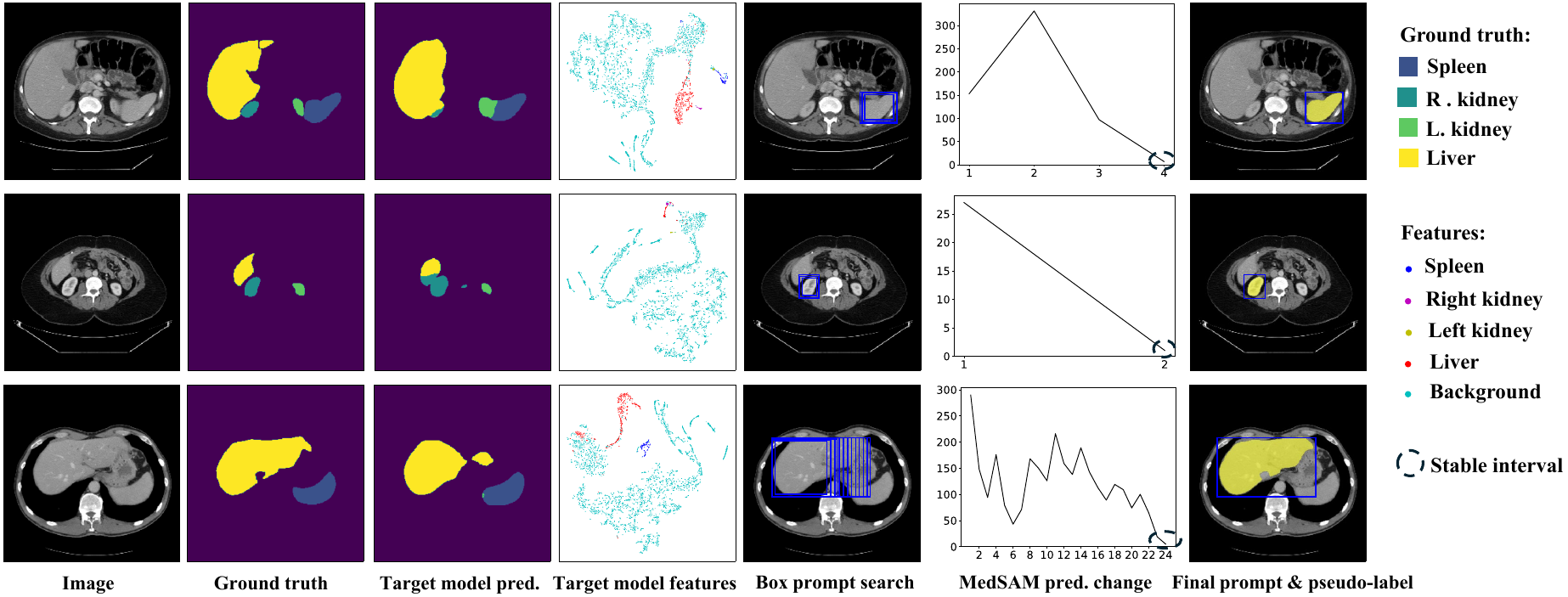}
\caption{Visualization of box prompt search and pseudo-labels. From left to right: Target images; Ground truth; The predictions of target model after feature aggregation; The features produced by target model after feature aggregation; The procedure of box prompt search, with the boxes changing from small to large (For clarity, only the first, the last and part of the other prompts are drawn); \revise{The number of pixels of changed MedSAM predictions when the box prompt is switched from the last to the current, with the horizontal axis being the index of the current prompt;} The final selected box prompt and the generated pseudo-labels.} \label{pl_ablation}
\end{figure*}

\subsubsection{Results for abdominal image segmentation}
For the abdominal datasets, we carry out experiments \revise{in two ``MRI to CT" scenarios and one ``CT to MRI" scenario}. The quantitative comparison results are shown in Table~\ref{quanti_result_abdominal1} \revise{and Table~\ref{quanti_result_abdominal2}}. It can be observed that there is a huge difference between the ``Source model" and the ``Target supervised", showing the large domain gap between the domains of different modalities.
All the SFDA methods surpass the source model. DPL\cite{dpl} and AdaMI\cite{adami} obtain moderate improvements compared to the source model. UPL\cite{upl} demonstrates excellent performance in its own experiments with relatively small domain gaps, but exhibits limited accuracy when facing a large domain gap. FVP\cite{fvp} and ProtoContra\cite{protocontra} obtain higher Dice values but are still not optimal.
Our proposed DFG yields the best performance with the highest average Dice ($84.9\%$, $85.2\%$, \revise{$85.7\%$}) and the lowest average ASSD (1.99, 1.00, \revise{2.27}).

The performance of naively adopting MedSAM is also analyzed. Directly obtaining box prompts according to the predictions of ProtoContra\cite{protocontra} and feeding them to MedSAM brings only slight boosts in Dice coefficients (74.4\%$\rightarrow$75.3\% and \revise{77.6\%$\rightarrow$81.9\%} for MRI$\rightarrow$CT) or even worsen ProtoContra's original results
(80.5\%$\rightarrow$75.0\% for CT$\rightarrow$MRI), as MedSAM requires a box prompt whose all four edges are close to the ground truth to produce a decent segmentation result. Existing SFDA methods adapt the source model to the target domain to some extent but are not specifically designed for producing a box prompt under the domain gap. This highlights our approach's significance in bridging the gap between SFDA and SAM.

The visualization of some segmentation results by different SFDA methods is shown in Fig.~\ref{quali}. Note that the results of our approach are more similar to the ground truth.

\subsubsection{Results for prostate image segmentation}
We further compare the performance of different SFDA methods on the prostate datasets, with NCI-ISBI and QUBIQ as the source and target domains, respectively.
Table~\ref{quanti_result_prostate} presents the quantitative evaluation results. The existing methods achieve Dice scores ranging from $73.6\%$ to $86.6\%$. Our DFG greatly improves the source model and yields the best performance with the highest Dice ($93.3\%$) and
the lowest ASSD (2.18), which are close to the results of ``Target supervised". The qualitative comparison is shown in the last two rows of Fig.~\ref{quali}.

\vspace{-0.7em}
\subsection{Ablation Study}
In this subsection, we conduct experiments to analyze the
effects of our designs. The results on the CHAOS to BTCV adaptation are reported.

\subsubsection{Ablation study of each component}
We conduct ablation experiments on each of the proposed components. The quantitative results are shown in Table~\ref{ablation_component}. Each component of our method lead to a performance improvement. Under a large domain gap between MRI and CT, using FA alone increases the average Dice to $73.2\%$, showing FA can preliminarily alleviate the domain shift. Further introducing TBS improves the Dice to $80.6\%$. This shows that TBS refines the pseudo-labels by propagating on the class-wise clustered features based on the aggregated target model feature space. Then we add the MBS which further varies the box prompt with the guidance of MedSAM features, which increases the Dice to $82.9\%$. Additionally incorporating the CP, corresponds to our proposed DFG approach, yielding the best Dice of $84.9\%$. We have also tried applying post-processing to the refined pseudo-labels obtained with only the guidance of target model features (i.e., removing MBS), and the Dice drops to $82.6\%$. Note that both methods with CP show significantly lower ASSD values, owing to the removed false positive regions.

\subsubsection{Effect of pseudo-label refinement and box prompt search}

\begin{table}[t]
\centering
\caption{Dice Scores of the Pseudo-labels on The Training Set at Different Phases of our approach.}\label{pl_ablation_quanti}
\setlength{\tabcolsep}{3pt}
\begin{tabular}{c|ccccc}

\toprule[1.5pt]

Methods & Liver & R. Kidney & L. Kidney & Spleen & Average \\
\hline

Before DBS& 84.3 & 67.1 & 68.7 & 71.6 & 72.9\\
After DBS & 91.4 & 75.8 & \textbf{76.3} & 79.6 & 80.7\\
After DBS \& CP & \textbf{91.9} & \textbf{80.9} & 76.1 & \textbf{82.3} & \textbf{82.8}\\

\bottomrule[1.5pt]

\end{tabular}
\end{table}

We further investigate our proposed pseudo-label refinement via box prompt search. Table~\ref{pl_ablation_quanti} shows the Dice scores at different phases of our approach. It can be observed that the quality of pseudo-labels are substantially improved by DBS and additionally advanced by CP. Three examples of pseudo-label refinement are provided in Fig.~\ref{pl_ablation}. For the image on the top, there exists confusion between the spleen and left kidney, and the prediction of spleen is smaller than its ground truth. Based on the preserved feature affinity property illustrated in Section~\ref{section3}, our proposed TBS gradually propagates on the class-wise clustered target model features of spleen. As a result, the box prompt stops expansion when it is close to the ground truth and the MedSAM output becomes stable. The right kidney of the image in the middle suffers from false positive predictions. A proper bounding box prompt is also found with the assistance of class-wise clustered target model features. The liver of the image at the bottom is troubled with a large false negative region, and faces the challenge of a more irregular shape. Nonetheless, relying on the property of MedSAM features, the box prompt is expanded in the correct direction, which leads to superb pseudo-labels.

\subsubsection{Comparison with finetuning Segment Anything Model}\label{ablation_finetune}

\begin{table}[t]
\centering
\caption{Comparison with Acquiring Pseudo-labels with Existing SFDA Method then Finetuing SAM. Dice Scores are listed.}\label{comparison_samed}
\setlength{\tabcolsep}{2pt}
\begin{tabular}{c|ccccc}

\toprule[1.5pt]

Methods & Liver & R. Kidney & L. Kidney & Spleen & Average \\
\hline

ProtoContra\cite{protocontra} & 85.4 & 75.5 & 68.0 & 68.9 & 74.4\\
ProtoContra+SAMed\cite{SAMed} & 79.0 & 60.9 & 59.6 & 69.6 & 67.3\\
Ours & \textbf{90.9} & \textbf{88.9} & \textbf{82.7} & \textbf{77.3} & \textbf{84.9}\\

\bottomrule[1.5pt]

\end{tabular}
\end{table}

\begin{table}[t]
\centering
\caption{\revise{Applying the Connectivity-based Post-processing also to the compared methods. Dice Scores are listed.}}\label{other_postprocess}
\setlength{\tabcolsep}{2pt}
\begin{tabular}{c|ccccc}

\toprule[1.5pt]

Methods & Liver & R. Kidney & L. Kidney & Spleen & Average \\
\hline
DPL\cite{dpl} & 39.0 & 35.3 & 58.6 & 42.4 & 43.9\\
DPL+CP & 35.7 & 31.9 & 58.8 & 46.0 & 43.1\\\hline
AdaMI\cite{adami} & 72.8 & 57.0 & 63.5 & 47.1 & 60.1\\
AdaMI+CP & 72.9 & 66.9 & 80.8 & 51.5 & 68.0 \\\hline
UPL\cite{upl} & 78.1 & 61.0 & 68.2 & 38.9 & 61.5\\
UPL+CP & 81.8 & 59.8 & 68.4 & 39.3 & 62.3\\\hline
ProtoContra\cite{protocontra} & 85.4 & 75.5 & 68.0 & 68.9 & 74.4\\
ProtoContra+CP & 85.5 & 77.3 & 71.4 & 68.9 & 75.8\\\hline
Ours w/o CP & \textbf{91.1} & 83.0 & 81.9 & 75.7 & 82.9 \\
Ours & 90.9 & \textbf{88.9} & \textbf{82.7} & \textbf{77.3} & \textbf{84.9}\\

\bottomrule[1.5pt]

\end{tabular}
\end{table}

\begin{figure}[t]
\centering
\includegraphics[width=\columnwidth]{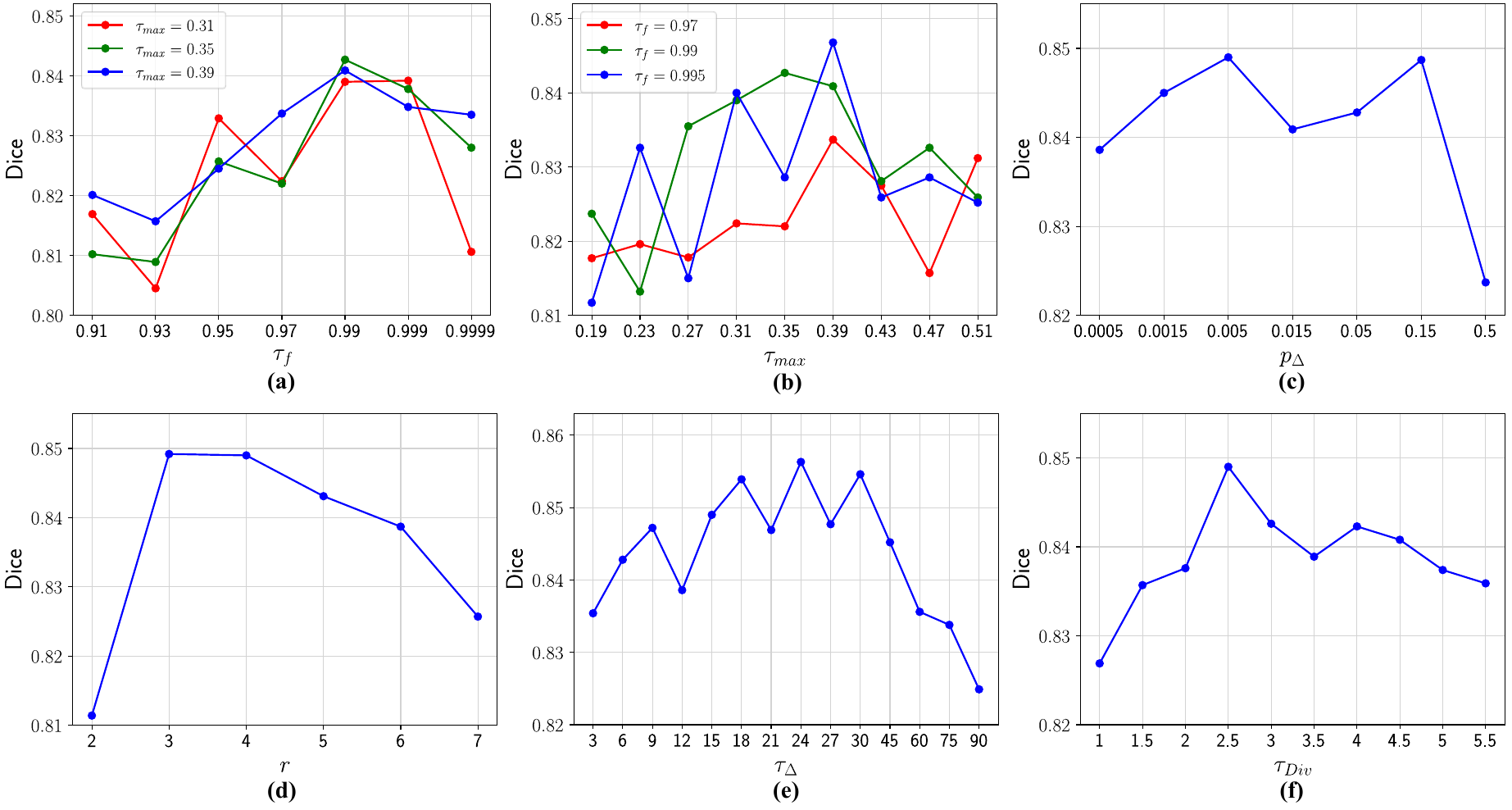}
\caption{\revise{Performance of our method with different hyper-parameter values. Dice scores are given.}} \label{param_ablation}
\end{figure}

An alternative way of leveraging SAM in SFDA is finetuning SAM to customize it for a specific task of medical image segmentation. To compare this method with our approach of leveraging SAM, we generate pseudo-labels with a recent SFDA method, ProtoContra\cite{protocontra}, then use the pseudo-labels to finetune SAM with the strategy in SAMed\cite{SAMed}. SAMed applies the low-rank-based finetuning method to the SAM encoder and finetunes it together with the prompt encoder and the mask decoder. We follow the pipelines and parameters in SAMed. The results are presented in Table~\ref{comparison_samed}. The average Dice of finetuning SAM with the pseudo-labels generated by ProtoContra is even lower than that of ProtoContra. It is mainly because finetuning SAM is normally conducted on labeled datasets. Finetuning SAM on bad pseudo-labels will not necessarily bring a segmentation performance increase.

\begin{figure}[t]
\centering
\includegraphics[width=0.94\columnwidth]{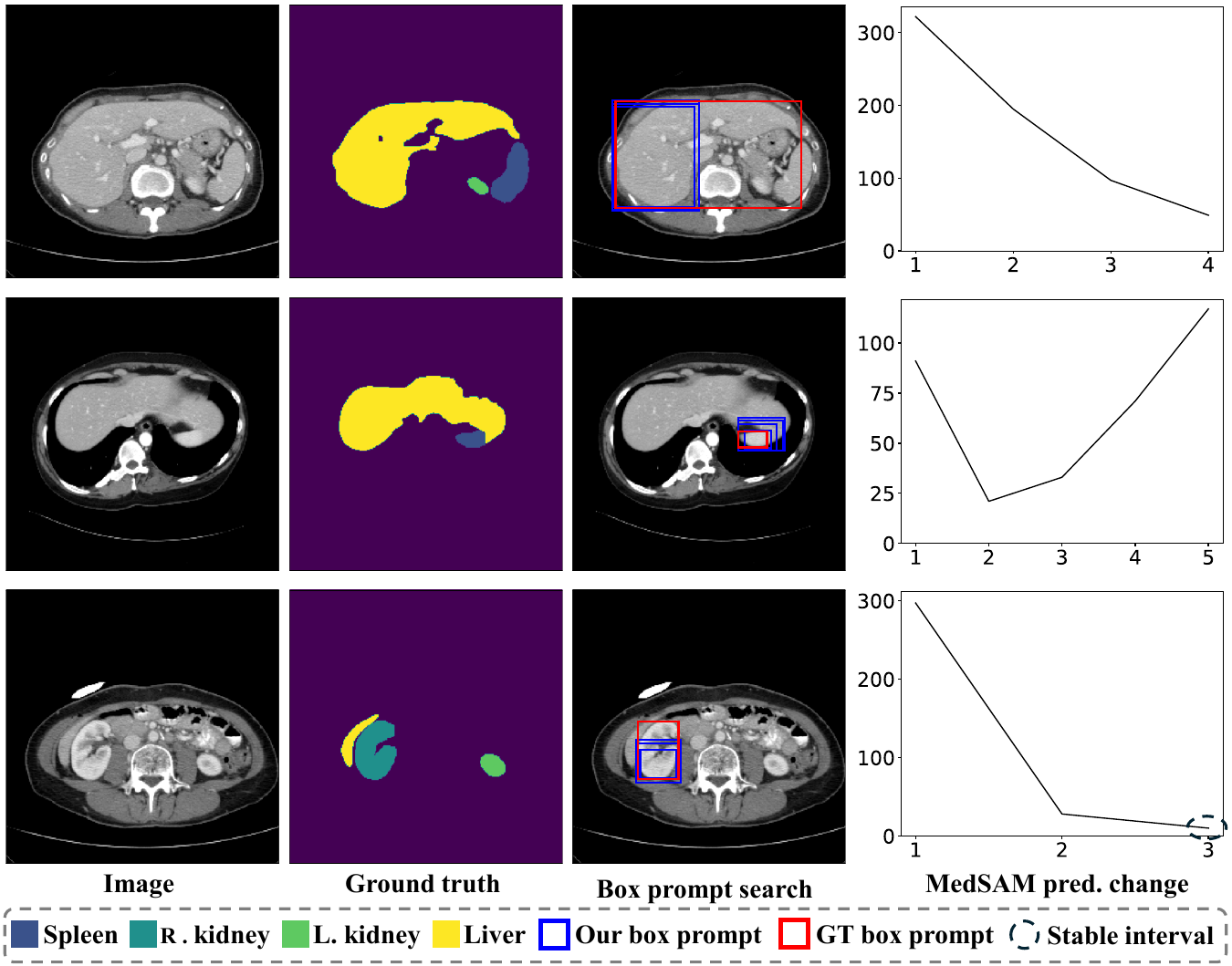}
\caption{\revise{Examples of failure cases of box prompt search. From left to right: Target images; Ground truth; Ground truth box prompt (red box), and box prompt search procedure (blue boxes), with boxes changing from small to large (For clarity, only the first, the last, and part of the other prompts are drawn); The number of pixels of changed MedSAM predictions when the box prompt is switched from the last to the current, with the horizontal axis being the index of the current prompt.}} \label{failure case}
\end{figure}

\begin{table*}[t]
\centering
\caption{\revise{The Running Time of Each Component in Our Method. For the Last Module, We Give the Results of Two Numbers of Epochs.}}\label{efficiency}
\begin{tabular}{c|c|c|c|c|c|c|c|c}

\toprule[1.5pt]
\multirow{2}{*}{Module} & \multirow{2}{*}{FA} & \multicolumn{4}{c|}{DBS}&
\multirow{2}{*}{CP} & \multicolumn{2}{c}{Training with $\mathcal{L}_{seg}$}\\
\cline{3-6}\cline{8-9}

& & \!SAM encoder\! & \!SAM decoder\! & \!Other\! & \!Total\! & & \!100 epochs (Final Dice: 84.9\%)\! & \!5 epochs (Final Dice: 83.9\%)\!\\\hline

Time (s) & 115.24 & 2438.47 & 166.82 & 429.97 & 3035.26 & 94.76 & 2984.19 & 149.21\\


\bottomrule[1.5pt]

\end{tabular}
\end{table*}

\subsubsection{Ablation study of hyper-parameters}

\revise{We investigate the effects of different hyper-parameters in our approach: the feature similarity threshold $\tau_f$ in TBS, the upper bound $\tau_{max}$ for the distance threshold in MBS, the $p_\Delta$ for selecting the initial pixel set, the $r$ for the propagation step size, the $\tau_\Delta$ for the criteria of a stable interval, and the distance threshold coefficient $\tau_{Div}$. The results are presented in Fig.~\ref{param_ablation}.} It can be seen that our method attains the best performance when the feature similarity threshold $\tau_f$ is 0.99 or 0.999. When $\tau_f$ is very low, the performance is bad because the TBS could propagate to another feature cluster. With $\tau_f$ increasing to 0.9999, our method suffers from performance degradation because TBS cannot propagate normally within a feature cluster. Besides, Fig.~\ref{param_ablation}(b) illustrates that the $\tau_{max}$ ranging between 0.31 and 0.39 yields the best results. Small $\tau_{max}$ hinders the propagation in MBS, whereas too large $\tau_{max}$ makes MBS cannot well exclude background pixels. \revise{Fig.~\ref{param_ablation}(c) demonstrates that the optimal value for $p_\Delta$ is 0.005. Regarding the propagation step size $r$, a too-small $r$ results in less significant box prompt changes, making it harder to distinguish between stable interval and unstable interval from the MedSAM output change. An overly large $r$ increases the risk of skipping the available stable interval. Our approach performs the best when $r$ is 3 or 4. In Fig.~\ref{param_ablation}(e), $\tau_\Delta$  denotes the threshold for a single step, with thresholds for two and three steps being $2\tau_\Delta$ and $3\tau_\Delta$, respectively. The parameter $\tau_\Delta$ ranging between 15 and 30 achieves the best results. A smaller $\tau_\Delta$ makes the stable interval detection prone to false negatives, while a larger $\tau_\Delta$ raises the likelihood of false positives. Finally, Fig.~\ref{param_ablation}(f) indicates that $\tau_{Div}=2.5$ is optimal. The trend in Fig.~\ref{param_ablation}(f) is governed by the same principle as in Fig.~\ref{param_ablation}(b). However, the performance degradation is less pronounced as $\tau_{Div}$ increases, thanks to the upper bound $\tau_{max}$ preventing the distance threshold from increasing excessively.}

\subsubsection{\revise{Applying the connectivity-based post-processing to other methods}}

\revise{
We also apply the introduced connectivity-based post-processing (CP) in Section~\ref{cp} to the other SFDA methods, with the results presented in Table~\ref{other_postprocess}. Generally, the CP, which preserves only the largest connected component for each class, can either enhance segmentation results (e.g., AdaMI, UPL, and ProtoContra) or degrade it (e.g., DPL), because the removed isolated smaller components can be either false positives or true positives depending on the property of the model outputs. In addition, applying CP typically results in marginal improvements except for AdaMI. In contrast, the CP tailored to the specific properties of the proposed DBS yields a more significant performance gain to our method. This is because DBS will start the box prompt search once the target model determines a class k exists in a 2D slice. Even only a small region of an organ belonging to the background is misclassified as class k, it might cause the entire organ in the background to be segmented as class k. The CP can effectively remove such increased false positive regions that belong to another organ disconnected from class k, thus making it complement DBS. Moreover, our approach still demonstrates superior Dice scores compared to other SFDA methods combined with CP.}

\subsubsection{\revise{Analysis of efficiency and parameters}}

\revise{Table~\ref{efficiency} shows the time consumption of our framework's each component. While our box prompt search involves iteratively expanding box prompts and computing MedSAM outputs, the inference on MedSAM encoder is performed just once for each image to extract an image embedding, which is then repetitively used for the inference of MedSAM decoder with varying prompts. Consequently, the box prompt search brings minimal overhead. Additionally, the training process with post-processed pseudo-labels is set to 100 epochs to ensure a full convergence. However, as shown in Table~\ref{efficiency}, this process can be reduced to 5 epochs with only a $1\%$ performance loss. The MedSAM encoder inference accounts for a majority of the total runtime, which can be alleviated by using the newly released lightweight version of MedSAM. Overall, our algorithm is computationally efficient and does not introduce much additional overhead from incorporating an extra foundation model.}

\revise{In terms of the number of parameters, DFG freezes MedSAM and trains only the target model without introducing additional trainable networks for adaptation. Namely, the trainable parameters are limited to the target segmentation network, which are the same as or fewer than previous SFDA methods, with the exception of FVP, which solely trains a visual prompt. Despite this, our method surpasses FVP in performance.}

\subsubsection{\revise{Failure case analysis}}

\revise{Although the proposed DFG achieves overall superior SFDA results, our box prompt search still fails to find satisfactory box prompts in some cases, some examples of which are illustrated in Fig.~\ref{failure case}. The liver in the top row occupies a large area, making its features more variable. Consequently, our box prompt search cannot identify the features of some portion of the liver as sufficiently similar to the prototype, resulting in this portion being excluded from the box prompt. In the middle row, the box prompt search does not manage to stop when the prompt varies near the ground truth prompt of the spleen, because MedSAM does not produce stable enough outputs during this phase. On the other hand, the box prompt in the bottom row prematurely stops expanding when it encounters an accidental stable interval in MedSAM’s output. A more robust version of SAM could mitigate the issues in the second and third cases. Besides, devising more reliable criteria for determining whether a pixel should be included in the box prompt, as well as a more sophisticated design for identifying stable intervals, could be directions for future work.}

\section{Conclusion}
\revise{Conventional SFDA algorithms are constrained by the limited information available from the source model and target data, hindering the adaptation performance from being further enhanced. Leveraging pre-trained foundation models offers a promising solution to address such a bottleneck.} 
In this paper, we propose a novel framework DFG for SFDA of medical image segmentation, which leverages the rich and generic knowledge of SAM. An auto-prompting strategy is designed with the guidance of target model features and SAM features. Experimental results on cross-modality abdominal datasets and multi-site prostate datasets show that our method outperforms existing SFDA methods.
\revise{In the future, one technical challenge of employing foundation models to assist SFDA of segmentation is the effective combination of the source model and the foundation model in the SFDA scenario. We build our framework based on the observed complementary characteristics of the source model features under domain shift and MedSAM features, which acts as an initial endeavor to fully integrate the heterogeneous knowledge of the two models and make use of their respective strengths. Additionally, our approach leverages both the MedSAM features and MedSAM segmentation output so as to utilize MedSAM’s knowledge to the fullest.  Hopefully, this can inspire further research that unlocks the foundation model's full potential to boost SFDA performance. We aspire for our innovations in SFDA to empower individual hospitals to handle data distribution shifts more effectively while preserving data privacy, and foster broader inter-institutional model sharing and collaboration.}

\bibliographystyle{IEEEtran}
\bibliography{reference.bib}{}

\end{document}